# SAR-RAG: ATR Visual Question Answering by Semantic Search, Retrieval, and MLLM Generation

David F. Ramirez*[ab], Tim Overman[b], Kristen Jaskie[b], Joe Marvin[b], Andreas Spanias[a]
[a]SenSIP Center, School of ECEE, Arizona State University, Tempe, Arizona, USA;
[b]Prime Solutions Group Inc, Goodyear, Arizona, USA

## ABSTRACT

We present a visual-context image-retrieval-augmented generation (ImageRAG)- assisted AI agent for automatic target recognition (ATR) of synthetic aperture radar (SAR) imagery. SAR is a remote sensing method used in defense and security applications to detect and monitor the positions of military vehicles, which may appear indistinguishable in images. Researchers have extensively studied SAR ATR to improve the differentiation and identification of vehicle types, characteristics, and measurements. Test examples can be compared with known vehicle target types to improve recognition tasks. New methods enhance the capabilities of neural networks, transformer attention, and multimodal large language models. An agentic AI method may be developed to utilize a defined set of tools, such as searching through a library of similar examples. Our proposed method, SAR Retrieval-Augmented Generation (SAR-RAG), combines a multimodal large language model (MLLM) with a vector database of semantic embeddings to support contextual search for image exemplars with known qualities. By recovering past image examples of known true target types, our SAR-RAG system can compare similar vehicle categories, thereby improving ATR prediction accuracy. We evaluate this through search and retrieval metrics, categorical classification accuracy, and numeric regression of vehicle dimensions. These metrics all show improvements when SAR-RAG is added to an MLLM baseline method as an attached ATR memory bank.

## 1. INTRODUCTION

Automatic target recognition (ATR) of vehicles in synthetic aperture radar (SAR) imagery is a challenging problem in remote sensing and defense. SAR-formatted images extrapolate spatial data from reflected radar energy, especially from metallic structures such as vehicles. A scarcity of high-quality annotated SAR research datasets has complicated this task [1-2]. While traditional SAR ATR pipelines built on machine learning [3], convolutional neural networks [4-5], and, more recently, transformer-based models [6-7] have made progress under controlled experimental conditions, they often fail to generalize across diverse operational environments. These limitations, driven by data imbalance and scene variability, constrain both the accuracy and interpretability of current systems [8-10]. Overcoming these challenges is essential in achieving robust, reliable, and explainable SAR ATR performance suitable for real-world military and intelligence applications.

Recent advances in multimodal large language models (MLLM) and multimodal learning have introduced retrieval-augmented generation (RAG) as a promising framework for addressing these limitations [11]. RAG enhances model reasoning by dynamically retrieving relevant knowledge—such as prior exemplars, contextual metadata, or semantic target descriptions—from an external database during inference. For SAR-based vehicle recognition, this matching to reference examples has well-documented benefits [12-13]: retrieved references can provide structural analogies, target-class priors, and spatial context that compensate for the scarcity and domain gaps in labeled SAR imagery. By fusing retrieved intelligence with SAR observations, our proposed RAG-driven systems can support more interpretable, data-efficient, and operationally robust ATR. This retrieval-augmented paradigm establishes a foundation for next-generation SAR recognition models that align machine perception with human language.

Recent research in remote sensing and geospatial analysis has begun to adopt RAG as an effective mechanism for integrating external knowledge into the interpretation of satellite imagery [14-16]. Traditional deep learning models have focused primarily on image classification and object detection in large-scale scenes, but these approaches often lack contextual awareness and generalization across environments. In contrast, RAG introduces a dynamic retrieval process that supplements imagery analysis with relevant textual, spatial, or multimodal information during inference.

*dframire@asu.edu; phone +1 623 853-0829; psg-inc.net

Three significant trends have emerged from this shift. Firstly, knowledge-grounded analysis enhances models' ability to reference prior examples or domain metadata [14], thereby improving interpretability and adaptability to new scenarios. Secondly, multimodal retrieval connects visual and linguistic information, facilitating richer semantic understanding for tasks such as captioning [15], question answering, and change detection. Lastly, spatially aware retrieval incorporates geographic structures and relationships into reasoning, enabling models to consider location-dependent context [16]. Collectively, these advancements indicate a move toward more intelligent, interpretable, and operationally robust geospatial AI systems.

### 1.1 Problem Statement

RAG introduces a robust framework for addressing the persistent limitations of SAR-based ATR. By integrating dynamic retrieval with generative reasoning, RAG enables models to augment learned SAR feature representations with externally sourced knowledge regarding observed targets. In our approach, a retrieval module queries a curated repository containing SAR exemplars, vehicle signatures, and contextual intelligence. At the same time, a generative reasoning component generates ATR predictions conditioned on both observed and retrieved evidence. This retrieval-guided mechanism grounds recognition in relevant prior knowledge, thereby mitigating hallucination, enhancing robustness in data-scarce or novel conditions, and improving interpretability.

While RAG introduces dynamic knowledge access into SAR-based ATR, its long-term effectiveness depends on the adaptability and continual relevance of its retrieval corpus and generative components. In real-world operational settings, new vehicle categories, sensors, and environmental conditions frequently emerge, rendering static models inadequate for sustained performance. To overcome these limitations, we propose integrating continual learning into the RAG framework to enable adaptive knowledge acquisition and incremental model refinement without catastrophic forgetting.

Our method, Synthetic Aperture Radar Retrieval-Augmented Generation (SAR-RAG), combines a multimodal large language model (MLLM) with a vector database of semantic embeddings to support contextual search for vehicle exemplars with known qualities. By combining retrieval and generation, our SAR-RAG system can achieve greater data efficiency, contextual awareness, and operational maintainability, marking a significant step toward explainable, mission-ready radar intelligence.

In our continuous RAG system, both the retrieval database and the generative reasoning model evolve over time. The retrieval index continually incorporates newly acquired SAR samples, contextual metadata, and mission-specific intelligence. At the same time, the generative module incrementally updates its internal representations via parameter-efficient, memory-constrained adaptation. This joint evolution allows the ATR system to preserve prior expertise while adapting to novel conditions. By coupling retrieval-augmented reasoning with continual learning, the proposed framework advances toward a form of sustainable adaptive intelligence that continuously integrates new radar observations, maintains accumulated knowledge, and enhances generalization across diverse and evolving operational domains.

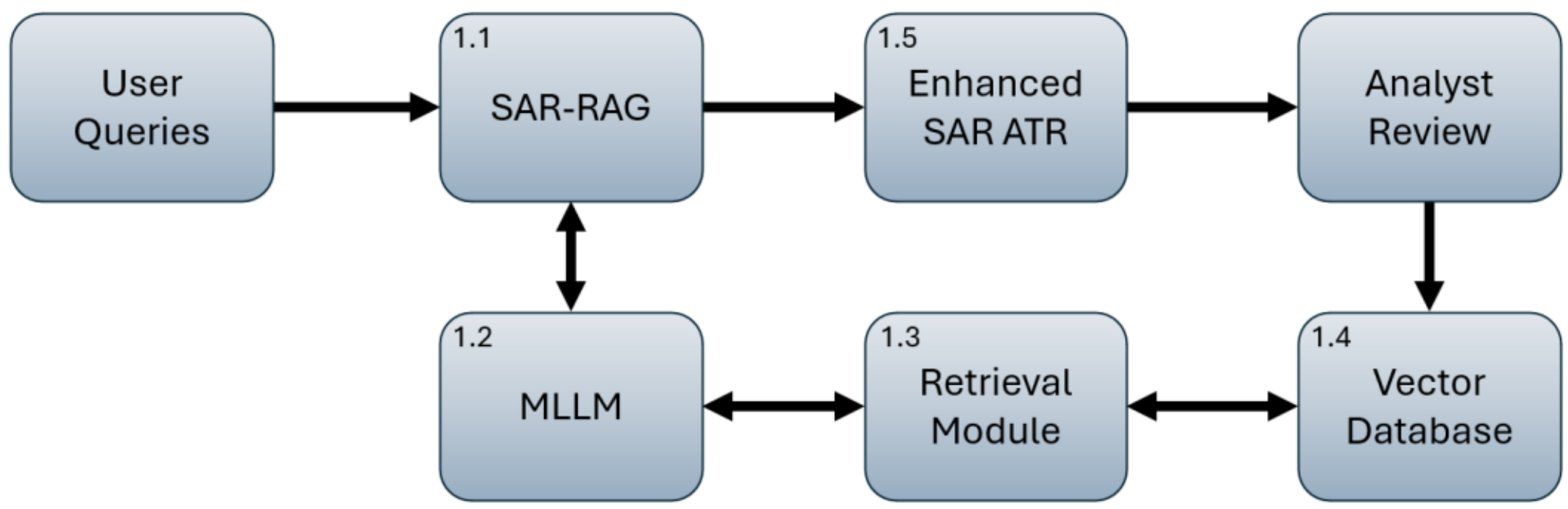

Figure 1. The SAR-RAG system diagram illustrates a multimodal large language model (MLLM) retrieving closely matching context from the vector database. A notional continual learning process includes a human-in-the-loop verifier.

## 2. BACKGROUND

### 2.1 Addressing Machine Learning Challenges

Large Language Models (LLM) have achieved exceptional performance in natural language understanding, reasoning, and generation. However, their knowledge remains inherently static and bounded by the information available at the time of machine learning training. This limitation poses a fundamental challenge for applications that require up-to-date, domain-specific, or verifiable information. Although LLMs generate coherent text, their reliance on fixed neural network parameters often yields responses that are outdated, incomplete, or factually imprecise in dynamic or specialized domains. Overcoming this constraint is essential for advancing toward more adaptive, trustworthy, and knowledge-aware language intelligence systems.

RAG overcomes this limitation by introducing a retrieval stage before generation, enabling models to dynamically access external databases, document collections, or knowledge graphs at inference time [11]. In this framework, retrieved information is fused with the model's internal representations, bridging the gap between parametric knowledge encoded in the model's parameters and non-parametric memory dynamically retrieved from up-to-date, external sources.

This integration enables the model to ground its reasoning in verifiable and contextually relevant information rather than relying solely on learned associations. As a result, RAG-enhanced systems produce outputs that are more factually accurate, contextually rich, and grounded in real-world data [11]. By coupling retrieval with generation, RAG transforms the LLM from a static text generator into an adaptive, evidence-based reasoning system capable of continuous knowledge integration and trustworthy language generation.

### 2.2 Multimodal Search, Retrieval, and Generation

Traditional RAG frameworks have primarily concentrated on text-based retrieval, enabling LLM to enhance their responses with externally sourced, up-to-date written information. While this approach has proven effective for factual reasoning, it does not account for the extensive knowledge embedded in visual modalities, such as imagery, diagrams, and formatting. As visual data becomes increasingly central to modern artificial intelligence workflows, the limitations of purely textual retrieval have become more apparent. Text-based RAG systems lack the capacity to interpret or generate responses grounded in visual evidence, constraining their ability to reason about spatial relationships, object structures, and scene-level context. Addressing this gap requires extending RAG to integrate visual information directly into the retrieval and generation process, thereby enabling multimodal reasoning that unifies language and perception.

ImageRAG [17] extends the traditional RAG framework by incorporating visual information as a central component of retrieval and reasoning [18-20]. These RAG systems index and retrieve visual content, such as images, maps, or remote-sensing scenes [14-16], alongside textual evidence to provide richer, multimodal context. Imagery is encoded as high-dimensional numeric embeddings using vision encoders or multimodal models and stored in vector databases, enabling efficient similarity-based retrieval.

During inference, the system retrieves visual instances that are semantically similar to the input query and integrates them into the generative reasoning process. This fusion enables the model to reason jointly across visual and textual modalities, interpret spatial relationships, and generate explanations grounded in observable evidence. By aligning visual retrieval with language generation, ImageRAG enhances factual grounding, contextual richness, and cross-modal understanding [17], thereby advancing toward truly evidence-based, visually grounded intelligence applicable to domains such as geospatial analysis, defense, and security.

### 2.3 Remote Sensing RAG

In the domain of remote sensing imagery and geospatial analysis, the adoption of RAG techniques is developing as a transformative trend. A key work is "RS-RAG: Bridging Remote Sensing Imagery and Comprehensive Knowledge with a Multimodal Dataset and Retrieval-Augmented Generation Model" [14]. This research introduces a multimodal knowledge database of high-resolution satellite imagery and extensive textual descriptions of over 14,000 landmarks globally. This data source is accessed by a retrieval-augmented vision-language model (VLM) for captioning, classification, and visual question answering (VQA) tasks. The authors report substantial gains over baseline VLM that lack retrieval components. Another published work [15] addresses the challenge of ultra-high-resolution (UHR) imagery by using a retrieval mechanism to select relevant image patches for analysis. This ImageRAG framework enables a model to focus on critical context within large scenes, thereby mitigating memory/compute issues while improving interpretive accuracy. Also concentrated on geospatial reasoning, the "Spatial-RAG: Spatial Retrieval Augmented Generation for Real-World Spatial Reasoning Questions" framework [16] extends RAG with a geospatial database, combining spatial location retrieval with semantic information to answer location-based queries using an LLM. RAG is evolving beyond purely image- or text-based retrieval to more structured spatial retrieval for geospatial tasks.

## 3. METHODS

The ImageRAG paradigm offers a novel approach to enhancing understanding and reasoning with SAR imagery. By integrating visual retrieval with generative reasoning, our SAR-RAG system identifies and retrieves pertinent SAR image exemplars, object templates, or scene segments from a curated database, then merges them with the current query context to enrich the model's reasoning. In the context of VQA, this framework enables an AI system to accurately answer textual queries about SAR scenes while grounding its responses in retrieved visual evidence from analogous imagery. Fig. 1 shows a system diagram with data-flow arrows and processing steps.

### 3.1 Data Corpus & Vector Database

The Moving and Stationary Target Acquisition and Recognition (MSTAR) dataset serves as a widely recognized benchmark for evaluating SAR–based ATR [1-2]. This comprises X-band SAR imagery of several military ground vehicles collected under controlled conditions with varying depression angles and aspect orientations. Each sample includes a greyscale SAR image chip, a complex-phase raster, and detailed metadata, including target class, serial number, collection angle, azimuth orientation, and configuration parameters. These characteristics make MSTAR particularly well-suited for constructing a retrieval-augmented corpus to support multimodal reasoning. We combine portions of the MSTAR Mixed Targets and T-72 Variants datasets. Specifically, we utilize the 2S1, BRDM-2, BTR-60, D7, SLICY, T-62, T-72, ZIL-131, and ZSU-23-4 vehicle types. This results in a total of 14,108 SAR image chips with metadata. Note that the T-72 vehicle type is overrepresented in the data, resulting in a 10:1 imbalance versus the least-represented target. We perform a 50/50 stratified random split of this data to form training and validation sets, repeating this process several times to form multiple experiments.

To develop the ImageRAG corpus, we process the MSTAR training subset target images to conform to a multimodal vector database. Each image chip, representing a specific vehicle and sensor configuration, is encoded with a standard embedding model utilizing the LlamaIndex framework [21]. The resulting embeddings are indexed in a Qdrant vector database [22] to enable similarity-based retrieval. In parallel, associated vehicle metadata is also stored in the Qdrant database, capturing written attributes such as target type, vehicle characteristics, measurements, depression angle, azimuth angle, and condition, with each metadata entry linking to its corresponding image via a unique identifier.

This hybrid architecture, combining visual embeddings and structured metadata, enables both content-based retrieval and context-aware filtering. During inference, when presented with a query image or a natural-language question, the system retrieves semantically similar SAR exemplars along with their contextual metadata to ground the model's reasoning. This integrated design supports downstream tasks such as VQA, target recognition, and explanation generation, thereby enhancing interpretability and ensuring that ground-truth SAR observations explicitly support system outputs.

### 3.2 Domain-Specific Cross-Modal Alignment

We use a variant of the Qwen2 vision-language transformer neural network [23] to generate image and language embeddings. This visual encoder transforms SAR imagery into semantic embeddings that preserve key structural and radar scattering characteristics relevant to retrieval and reasoning. Unlike optical imagery, SAR data exhibit unique properties such as speckle noise, anisotropic backscattering, and sensor-dependent distortions, which pose significant challenges for feature extraction. Consequently, vision encoders pretrained on natural images often perform suboptimally when applied directly to SAR data, necessitating domain-specific adaptation.

To address this, we fine-tune the image encoder on SAR data using supervised fine-tuning. This enables the model to learn class-discriminative radar features. Through this optimization, the encoder learns a radar-specific representation space that balances visual discrimination and semantic similarity. We follow model-training steps similar to our prior work described in "Towards a Large Language-Vision Question Answering Model for MSTAR Automatic Target Recognition" [6].

We train an MLLM with an attached SAR-RAG module for several VQA tasks. Similar to our prior work [6], we utilize the LLaVA-Next v1.6 Mistral-7B MLLM and parameter-efficient fine-tuning.

Table 1. Evaluation Results: Semantic Search and Retrieval Performance

<table>
<tr><th>Retrieval Metric</th><th>SAR-RAG</th><th>Random (class priors)</th></tr>
<tr><td><b>Accuracy@1-shot</b></td><td><b>77.72%</b></td><td rowspan="5">21.11%</td></tr>
<tr><td>Precision@2-shot</td><td>76.97%</td></tr>
<tr><td>Precision@3-shot</td><td>76.06%</td></tr>
<tr><td>Precision@4-shot</td><td>75.10%</td></tr>
<tr><td>Precision@5-shot</td><td>74.39%</td></tr>
<tr><td>Hit-Rate@2-shot</td><td>85.75%</td><td>37.76%</td></tr>
<tr><td>Hit-Rate@3-shot</td><td>88.90%</td><td>50.90%</td></tr>
<tr><td>Hit-Rate@4-shot</td><td>91.34%</td><td>61.26%</td></tr>
<tr><td>Hit-Rate@5-shot</td><td>93.54%</td><td>69.44%</td></tr>
<tr><td>Consensus@2-shot</td><td>68.19%</td><td>4.456%</td></tr>
<tr><td>Consensus@3-shot</td><td>61.58%</td><td>0.941%</td></tr>
<tr><td>Consensus@4-shot</td><td>57.48%</td><td>0.199%</td></tr>
<tr><td>Consensus@5-shot</td><td>53.54%</td><td>0.042%</td></tr>
<tr><td>Majority@3-shot</td><td>87.78%</td><td>11.49%</td></tr>
<tr><td><b>Majority@5-shot</b></td><td><b>92.41%</b></td><td>6.679%</td></tr>
</table>

### 3.3 Evaluation Metrics

We evaluate the performance of our proposed SAR-RAG system using a combination of retrieval, classification, VQA, and regression metrics to comprehensively assess accuracy and robustness. The SAR-RAG framework is compared with a baseline MLLM method trained similarly but without an ImageRAG component. These two methods are compared against the theoretical performance of arbitrary blind guesses weighted by the known distribution of vehicle types. This comparison sets a lower bound on performance for the complex evaluations that follow:

**3.3.1 Retrieval Performance**: Retrieval quality measures the system's ability to identify relevant SAR exemplars from the database given a query image. Given a validation image, we recover the five most similar image vectors using cosign similarity. The varieties of these five comparison vehicles are compared with the evaluation sample. The closest matching vector is evaluated on Accuracy. The Precision@N metric then measures the percentage of accurate retrievals among all five. The Hit-Rate@N measures whether any of the five are correct, while Consensus@N measures whether the samples are all correct. The MLLM baseline lacks a retrieval module and cannot be evaluated. Ultimately, the system uses a simple majority voting decision to determine the most probable target match.

**3.3.2 Automatic Target Recognition**: For classification-oriented evaluations, accuracy measures the correctness of target identification. We also evaluate the models' ability to generate a set of descriptive qualities associated with a vehicle, similar to our prior work [6]. We assess the model's ability to detect a mounted weapon system as a binary prediction problem.

**3.3.3 Quantitative VQA Metrics**: Natural language outputs generated by an MLLM can be challenging to evaluate. English grammar, word choice, semantic meaning, and factual correctness must be decoupled for the most precise evaluation. To focus on assessing factual correctness, we specifically assess the accuracy of numeric values. Continuous numeric attributes, such as the vehicle's weight, length, width, height, and weapon system size, are evaluated using Mean Absolute Error (MAE), Root Mean Squared Error (RMSE), and Mean Absolute Percentage Error (MAPE). The length estimates are measured in centimeters as they are consistently under a full meter. We explore this quantitative spatial reasoning in greater detail in our future work.

Table 2. Evaluation Results: Visual Question Answering

| **Evaluation Metric** | **SAR-RAG** | **Baseline MLLM [6]** | **Random (class priors)** |
|---|---|---|---|
| **Automatic Target Recognition (accuracy)** | | | |
| Vehicle Classification | **99.24%** | 99.04% | 20.95% |
| Armament Detection | 100.0% | 100.0% | 52.19% |
| Descriptive Qualities | **98.79%** | 98.47% | 37.27% |
| **Quantitative Physical Property Estimation** | | | |
| **Vehicle Weight** (metric tons) | | | |
| MAE | 0.428 | 0.530 | 15.00 |
| RMSE | 0.891 | 1.124 | 21.38 |
| MAPE | **1.24%** | 1.57% | 66.91% |
| **Vehicle Length** (centimeters) | | | |
| MAE | 7.926 | 10.00 | 161.4 |
| RMSE | 21.63 | 27.28 | 232.5 |
| MAPE | **1.46%** | 1.84% | 38.07% |
| **Vehicle Width** (centimeters) | | | |
| MAE | 5.582 | 7.041 | 51.79 |
| RMSE | 19.67 | 24.81 | 66.74 |
| MAPE | **1.92%** | 2.42% | 17.98% |
| **Vehicle Height** (centimeters) | | | |
| MAE | 79.87 | 100.7 | 44.21 |
| RMSE | 175.8 | 221.7 | 59.20 |
| MAPE | **35.65%** | 44.96% | 20.44% |
| **Armament Length** (centimeters) | | | |
| MAE | 12.18 | 15.36 | 307.8 |
| RMSE | 49.52 | 62.45 | 407.2 |
| MAPE | **2.55%** | 3.21% | 45.93% |

## 4. RESULTS

Tables 1 and 2 below show the performance of SAR-RAG on database-lookup matching and across VQA benchmarks.

The ImageRAG retrieval module was evaluated as likely to return examples that match the vehicle type. Despite the visual greyscale similarity among SAR images, the embedding method produced vectors that clustered similar vehicle types. Note how often all five vectors match a given query. This is evident in the t-distributed Stochastic Neighbor Embedding (t-SNE) vector dimensionality reduction algorithm and visualization in Fig. 2. Also of note is the separation of the SLICY target from the true vehicles. It is well documented that his radar reflection simulation target appears very different from the others, as shown in the vector embedding.

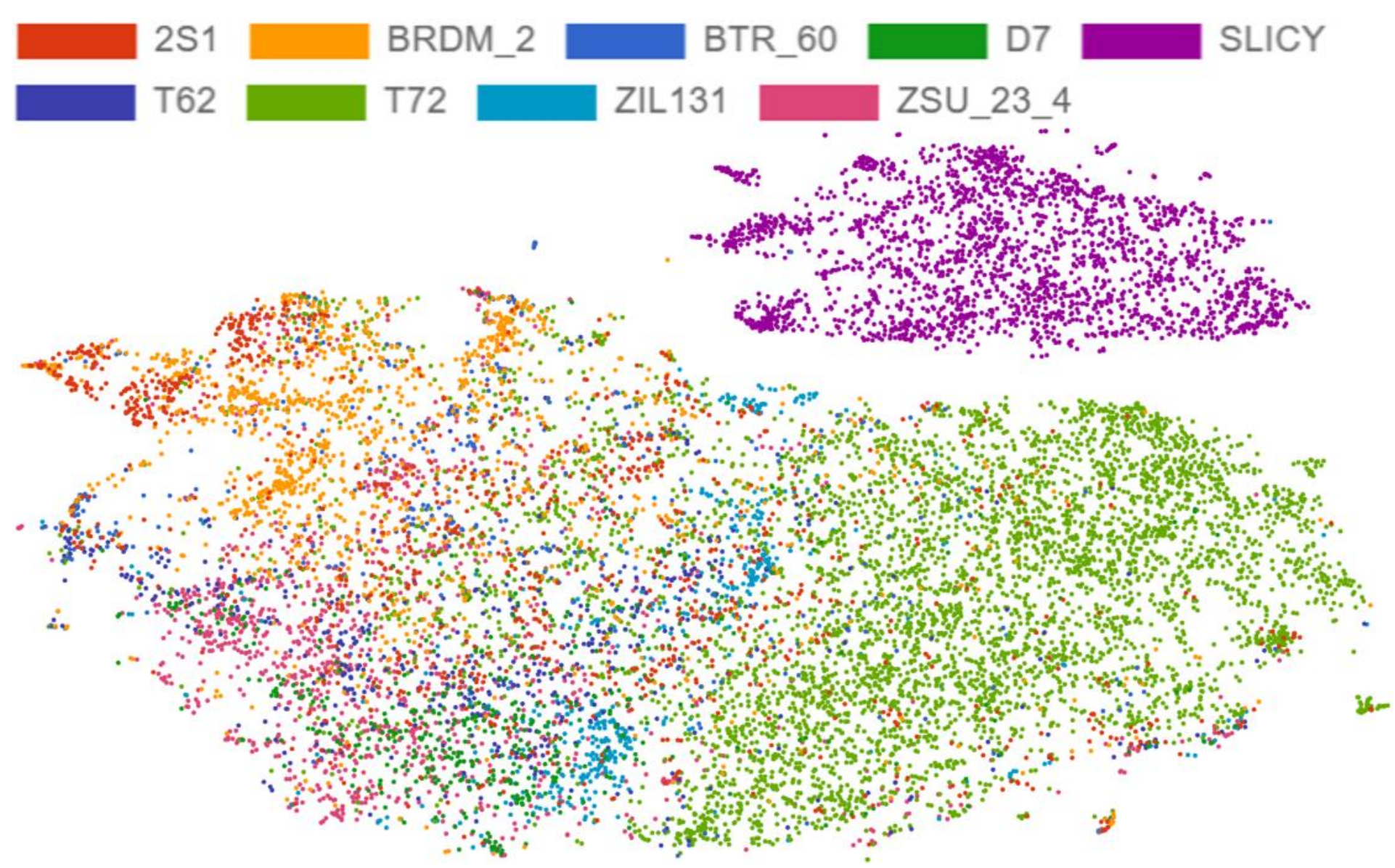

Figure 2. The t-SNE algorithm visualizes clusters of similar SAR image embeddings across various vehicle types.

The overall accuracy of the retrieval rates positively influences the generation of correct answers in subsequent evaluations, as the ground-truth data provides context for the MLLM's decision-making. This is most evident in the quantitative reasoning metrics. The SAR-RAG enhanced method is trained to consider similar vehicles and their known accurate measurements. This reduces the chances of serious hallucination, which we have documented as an issue in the baseline model. When the predicted value is erroneously off by an order of magnitude, this greatly inflates the error metrics. Providing a readily accessible source of information for the MLLM significantly reduces the likelihood of substantial error.

Finally, we observe minor improvements in the ATR classification metrics for the SAR-RAG-enhanced method relative to the baseline. Vehicle type prediction is slightly improved on average, although it is already very accurate. The set of descriptive qualities can be predicted precisely if the actual vehicle type is known. Without first inferring the vehicle type, generating all the vehicle's qualities is more likely to be error-prone. Similarly, the SAR-RAG retrieval is not perfect and can introduce errors if the incorrect vectors match.

## 5. CONCLUSIONS

SAR-RAG is a new technique for referencing a database of ATR targets before an MLLM predicts the answer for a VQA task. This additional information improves prediction performance across all benchmark tasks tested. We evaluated the retrieval efficacy of this vector search by retrieving the five nearest embeddings to a query image using cosine similarity. Our trained MLLM improved slightly with our proposed ImageRAG system, reaching nearly perfect accuracy. Finally, we benchmarked against several quantitative VQA tasks that infer the size of the target vehicles. This was greatly improved by the inclusion of the SAR-RAG system, which reduced rare but persistent significant errors in the baseline model.

In addition to improved accuracy, a RAG system offers other benefits. RAG reduces the amount of retraining required for transformer methods, since updates can be applied to the knowledge database rather than costly AI retraining. Additionally, filtering mechanisms can be added to the vector lookup to focus on specific environmental conditions, for example, retrieving only SAR images similar to a particular sensor elevation angle. RAG can also be extended to use more advanced database varieties. Graph databases or knowledge graphs [24] can be structured for RAG, unlocking even greater context discovery by the MLLM AI agent.

SAR-RAG is another compelling extension of MLLM methods and VQA tasks. Our work to date has extensively used the MSTAR and SAMPLE datasets. We have extended these traditional SAR ATR datasets with modern VQA text sequences to train MLLM for novel tasks. Our future work explores additional quantitative ATR VQA tasks and new data sources [25]. Researchers should consider what other application domains can benefit from powerful new AI models tailored to specific use cases.

## ACKNOWLEDGEMENTS

This work is supported by Prime Solutions Group Inc. and the ASU Sensor, Signal, and Information Processing (SenSIP) Center. A.I. tools were used during our literature review, proofreading, and editing, including AI2 Astra, ChatGPT, Google Gemini, and Grammarly.